\pdfoutput=1
\documentclass[sigconf]{acmart}
\usepackage{multirow}
\usepackage{subfig}
\usepackage{graphicx}
\usepackage{hyperref}
\usepackage{float}

\renewcommand\footnotetextcopyrightpermission[1]{} 

\begin{document}

\title[PBIC]{Patch-wise Features for Blur Image Classification}

\author{Sri Charan Kattamuru}
\affiliation{
  \institution{Applied Research, Swiggy}
  \city{Bangalore}
  \country{India}
}
\email{sricharan.k@swiggy.in}

\author{Kshitij Agrawal}
\affiliation{
  \institution{Applied Research, Swiggy}
  \city{Bangalore}
  \country{India}
}
\email{kshitij.agrawal@swiggy.in}

\author{Shyam Prasad Adhikari}
\affiliation{
  \institution{Applied Research, Swiggy}
  \city{Bangalore}
  \country{India}
}
\email{shyam.pa@swiggy.in}

\author{Abhishek Bose}
\affiliation{
  \institution{Applied Research, Swiggy}
  \city{Bangalore}
  \country{India}
}
\email{abhishek.bose@swiggy.in}

\author{Hemant Misra}
\affiliation{
  \institution{Applied Research, Swiggy}
  \city{Bangalore}
  \country{India}
}
\email{hemant.misra@swiggy.in}

\renewcommand{\shortauthors}{Kattamuru, et al.}

\begin{abstract}
  Images captured through smartphone cameras often suffer from degradation, blur being one of the major ones, posing a challenge in processing these images for downstream tasks. In this paper we propose low-compute lightweight patch-wise features for image quality assessment. Using our method we can discriminate between blur vs sharp image degradation. To this end, we train a decision-tree-based XGBoost model on various intuitive image features like gray level variance, first and second order gradients, texture features like local binary patterns. Experiments conducted on an open dataset show that the proposed low compute method results in 90.1\% mean accuracy on the validation set, which is comparable to the accuracy of a compute-intensive VGG16 network with 94\% mean accuracy fine-tuned to this task. To demonstrate the generalizability of our proposed features and model we test the model on BHBID dataset and an internal dataset where we attain accuracy of 98\% and 91\%, respectively. The proposed method is 10x faster than the VGG16 based model on CPU and scales linearly to the input image size making it suitable to be implemented on low compute edge devices.
\end{abstract}

\maketitle
\pagestyle{plain} 

\section{Introduction}
Images have become ubiquitous with the advent of smartphone devices with advanced camera systems. This explosion in digital image data has been the driving force behind many computer vision applications such as object detection, face recognition~\cite{masi2018deep}, medical image classification~\cite{basha2018rccnet}, document recognition, and self-driving cars~\cite{bojarski2016end}. These tasks often rely on high quality images with objects in sharp focus and any degradation in image quality leads to adverse performance~\cite{vasiljevic2016examining}.\par

Blur is one such undesirable degradation effect that is commonly found in images. This is caused by factors such as lack of focus or due to relative motion between the camera and target. Intuitively, we observe that the degree of blur in an image inversely effects the amount of information within an image. And this poses a challenge to extract semantic level information from it. Thus, blur classification is proposed as a preprocessing step to identify and reject low quality images either at the time of image acquisition or before processing them for downstream tasks.  

Blur detection, segmentation and image deblurring are areas of active research~\cite{cun2020defocus, shi2014discriminative, kupyn2019deblurgan} which help identify and rectify the images affected with blur. The objective of blur detection is to identify the region within an image that is affected with blur. While blur segmentation is posed as a pixel-wise classification task to generate a blur map. Image deblurring restores a sharp image from a given blurred image. Some approaches to image deblurring pose this task as an image filtering problem~\cite{sankhe2011deblurring}, while some approaches pose this as an image-to-image translation problem \cite{kupyn2019deblurgan}. \par

\begin{figure*}[t!]
\centering
\begin{tabular}{cc|cc|c|c}

\subfloat{\includegraphics[width=0.14\linewidth]{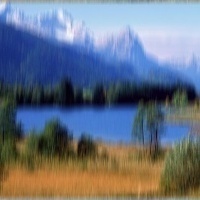}} &
\subfloat{\includegraphics[width=0.14\linewidth]{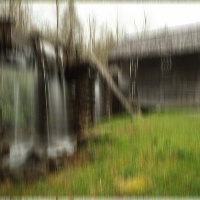}} &
\subfloat{\includegraphics[width=0.14\linewidth]{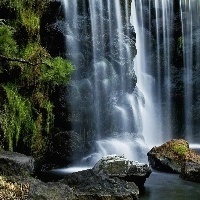}} &
\subfloat{\includegraphics[width=0.14\linewidth]{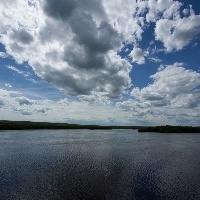}} &
\subfloat{\includegraphics[width=0.14\linewidth]{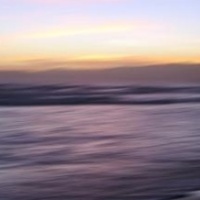}} &
\subfloat{\includegraphics[width=0.14\linewidth]{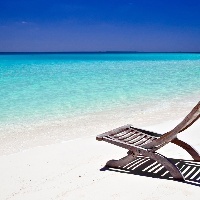}}
\\
\subfloat{\includegraphics[width=0.14\linewidth]{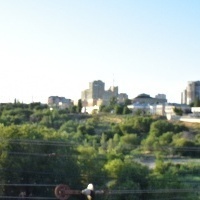}} &
\subfloat{\includegraphics[width=0.14\linewidth]{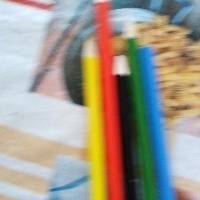}} &
\subfloat{\includegraphics[width=0.14\linewidth]{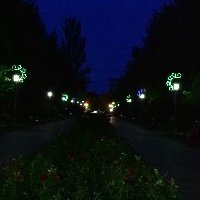}} &
\subfloat{\includegraphics[width=0.14\linewidth]{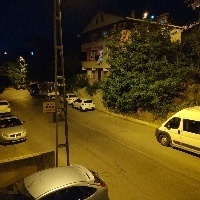}} &
\subfloat{\includegraphics[width=0.14\linewidth]{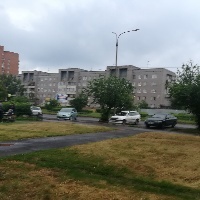}} &
\subfloat{\includegraphics[width=0.14\linewidth]{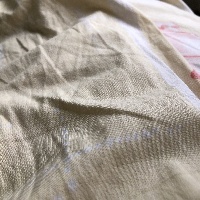}}
\\
\subfloat{\includegraphics[width=0.14\linewidth]{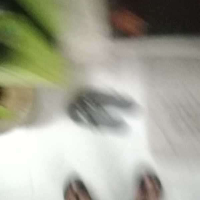}} &
\subfloat{\includegraphics[width=0.14\linewidth]{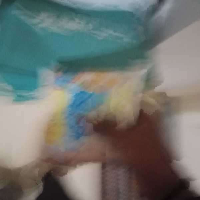}} &
\subfloat{\includegraphics[width=0.14\linewidth]{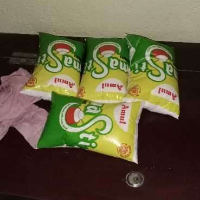}} &
\subfloat{\includegraphics[width=0.14\linewidth]{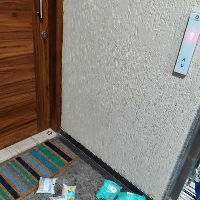}} &
\subfloat{\includegraphics[width=0.14\linewidth]{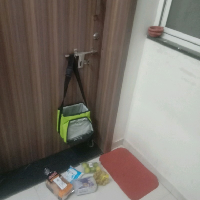}} &
\subfloat{\includegraphics[width=0.14\linewidth]{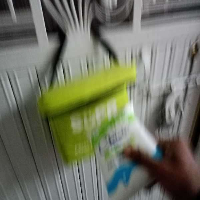}}
\end{tabular}

\caption{Figure shows the predictions by our best performing model, the rows correspond to samples from BHBID, KBD, Internal Dataset respectively. Columns 1 \& 2 correspond to the true positives of blur, columns 3 \& 4 correspond to the true positives in sharp, column 5 \& 6 correspond to the false positives in blur and sharp respectively}
\label{fig:prediction-samples}
\end{figure*}

The tasks of image quality assessment and blur classification can be divided into reference-based and non-reference based methods. Non-reference based methods are more challenging since we do not have any access to a sharp reference image which contains accurate information of the scene.\par

Convolutional neural networks (CNNs) are a well-known approach~\cite{he2016deep, krizhevsky2012imagenet} for image classification, but there exist a few obvious downsides of using CNNs for the task of blur classification. These are data-hungry models and tend to perform best with training on large scale datasets. They have parameters in the order of millions and are trained on optimized GPU-based architectures that take considerably long time to train. Even after training such a large model, in most cases, they cannot be deployed directly on the edge or low compute devices. We generally require sophisticated hardware designed specifically to run these models along with additional optimization steps like distillation, pruning or quantization ~\cite{liang2021pruning}. 

For our application we perform blur classification for image quality assessment, to identify and reject images that contain unintentional blur (as shown in Fig. \ref{fig:prediction-samples}). Additionally, we require a model that accurately performs this classification with fast inference times. In this work, 
\begin{itemize}
    \item We propose a method for non-reference based blur classification that utilizes conceptually intuitive features
    \item We train the proposed set of features using Extreme Gradient Boosting (XGBoost) classifier
    \item We train convolutional neural networks of varying complexity on the task of blur classification and compare these against our hand-crafted features across metrics like accuracy, roc-auc and benchmark them on inference time
\end{itemize}

We experimentally demonstrate that our proposed feature sets combined with XGBoost classifier present a model that is accurate and provides fast inference times.

\section{Related Works}

We focus on blur detection or classification at the image level to classify the presence of blur in an image. 

\textbf{Feature based methods -} Blur classification is a well studied area of research and multiple features like Statistical Features, Texture Features, Image quality metrics, Spectrum and Transform features, and Local power spectrum Features have been used to classify blur \cite{shi2014discriminative, wang2019automatic, ali2018analysis}. Gradient based methods like Laplacian and Tenengrad capture the regions of rapid changes in intensity, like edges. The presence of high level of edges can be correlated to sharpness of an image. This is extended to capture variance and mean of a laplacian at a global level to represent the amount of blur within an image~\cite{szandala2020convolutional}. The authors of \cite{wang2019automatic} use a combination of 35 features to classify different types of blur within an image by training a SVM in an one against one method and ensemble multiple such SVMs together to classify the type of blur. 

Global features do not present an accurate representation of blur around the main object of interest and in \cite{shi2014discriminative} the authors propose using both spatial and frequency features to train a naive bayes classifier on local patches. They further extend the concept to multiscale features to create a blur segmentation map. 
From \cite{ali2018analysis} it is observed that frequency domain features are less robust to noise. Hence, in our work, we draw inspiration from \cite{shi2014discriminative} and explore the use of gradient based operations applied at a patch level to images for robust blur classification. 

Local binary pattern (LBP) based descriptors have been used to give a robust representation of image texture. In \cite{ojala2002multiresolution} the authors propose a modification to LBP by using only uniform patterns, where the local grid contains only two transitions from zero to one or vice versa. The authors utilize this formulation for the task of defocus blur segmentation. 

\textbf{Deep Learning based methods -} Some work has been done to investigate the usefulness of CNNs as a feature extractor and in \cite{szandala2020convolutional} they replace handcrafted features with a convolutional neural network, however in their comparison the laplacian features perform better. Modern research in deep learning methods has focused on blur segmentation and rectification of various types of blur (defocus, motion, haze etc). 

Most of the approaches first detect the focused and out-of-focus regions within an image and subsequently use these segmented regions for deblurring. In \cite{cun2020defocus} the authors use VGG based FCN  to extract relevant features from blurred and sharp regions within an image. The benefit of this approach is that it bypasses the need for handcrafted features, while it also comes with a penalty of high computational requirement, limiting the use in edge devices and fast inference situation. In our work we adapt the feature extraction pipeline and fine-tune this for the task of blur classification. 

\begin{figure*}[ht]
\resizebox{0.9\textwidth}{!}{
\begin{tabular}{cccc}

\subfloat[Original image]{\includegraphics[width = 0.225\textwidth]{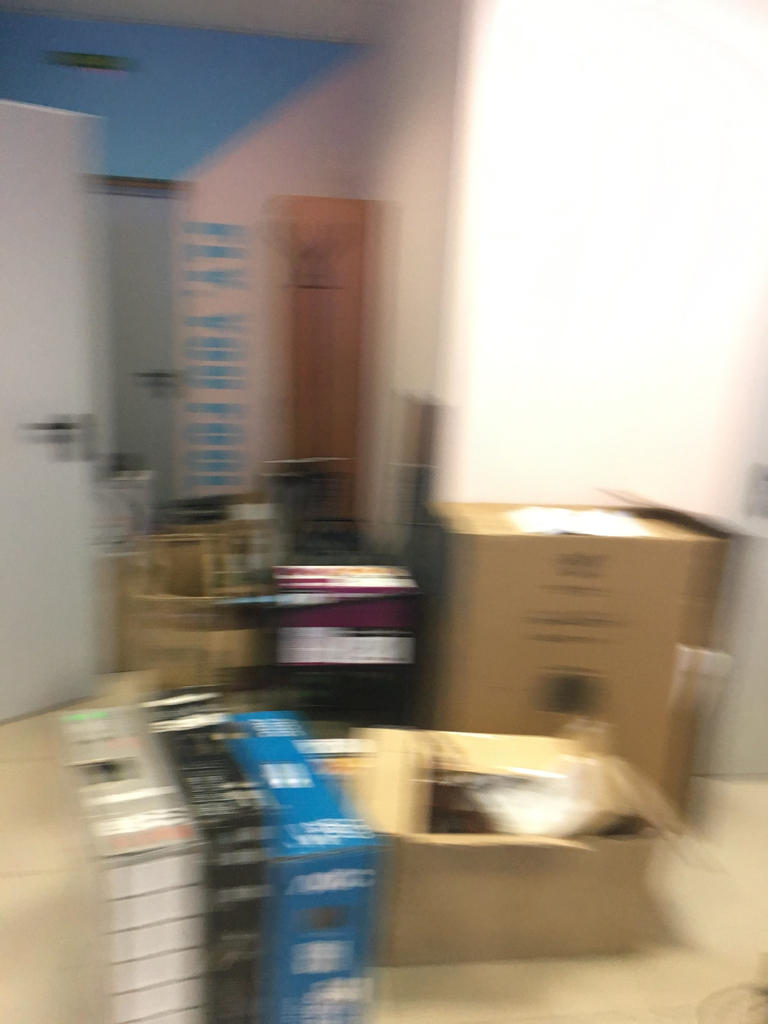}\label{original}} &
\subfloat[Image with text]{\includegraphics[width = 0.225\textwidth]{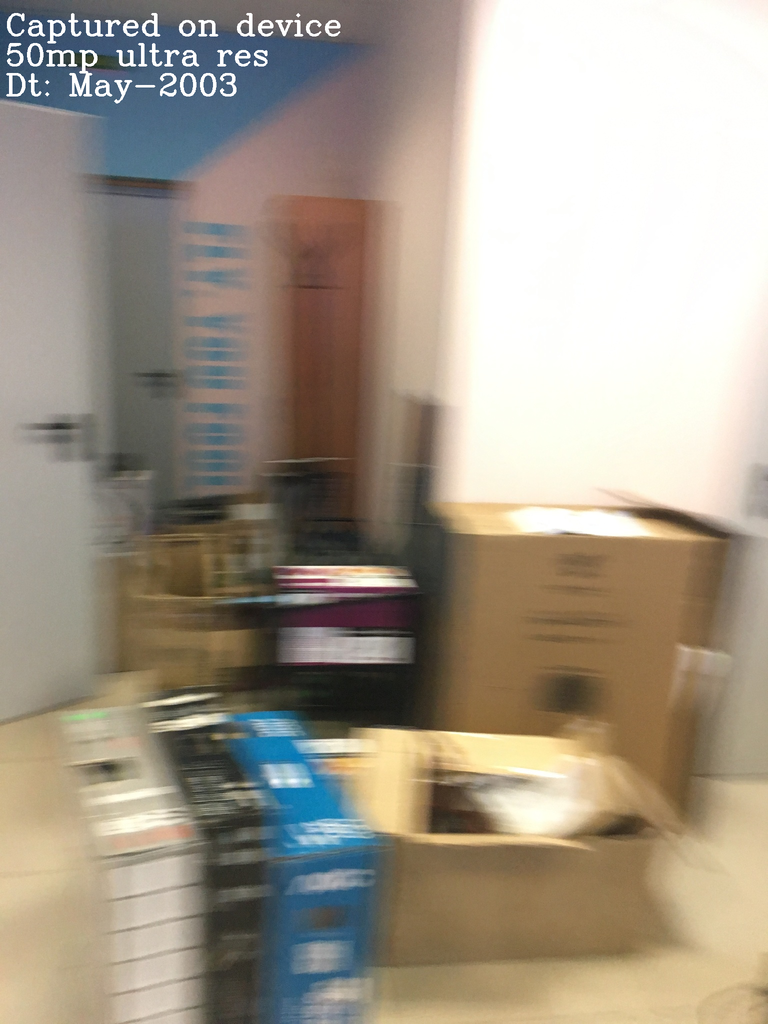}\label{watermark}}
&
\subfloat[Grid-wise inference on \textit{(a)} ]{\includegraphics[width=0.225\textwidth]{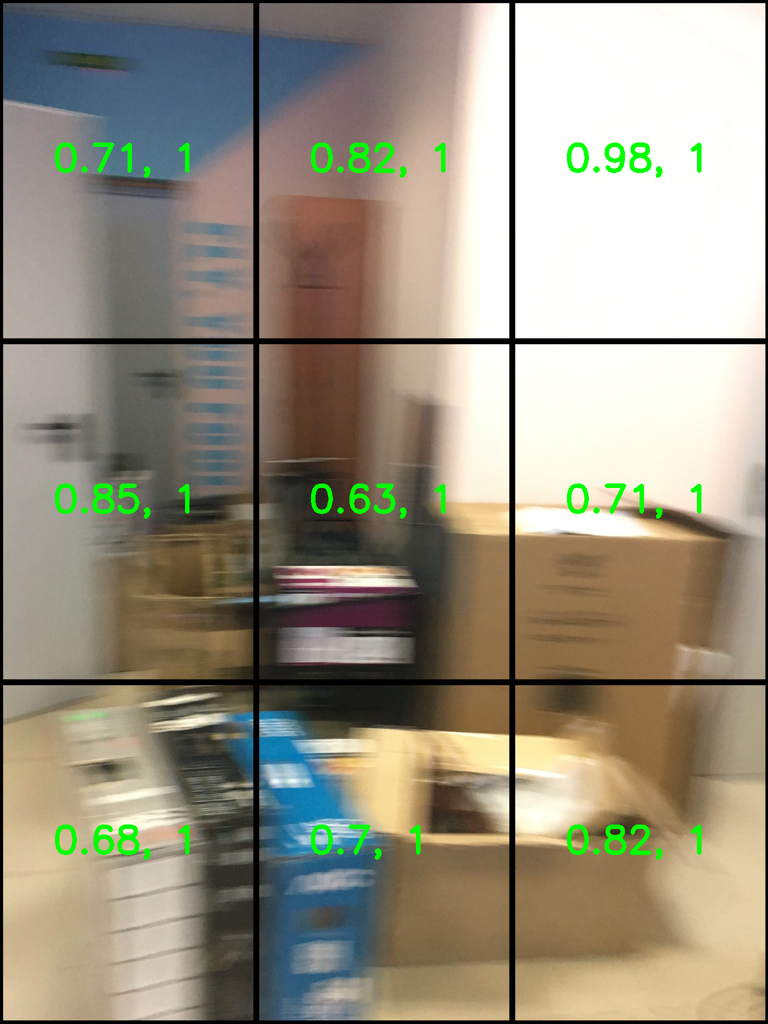}\label{original-grid}}
&
\subfloat[Grid-wise inference on \textit{(b)}]{\includegraphics[width= 0.225\textwidth]{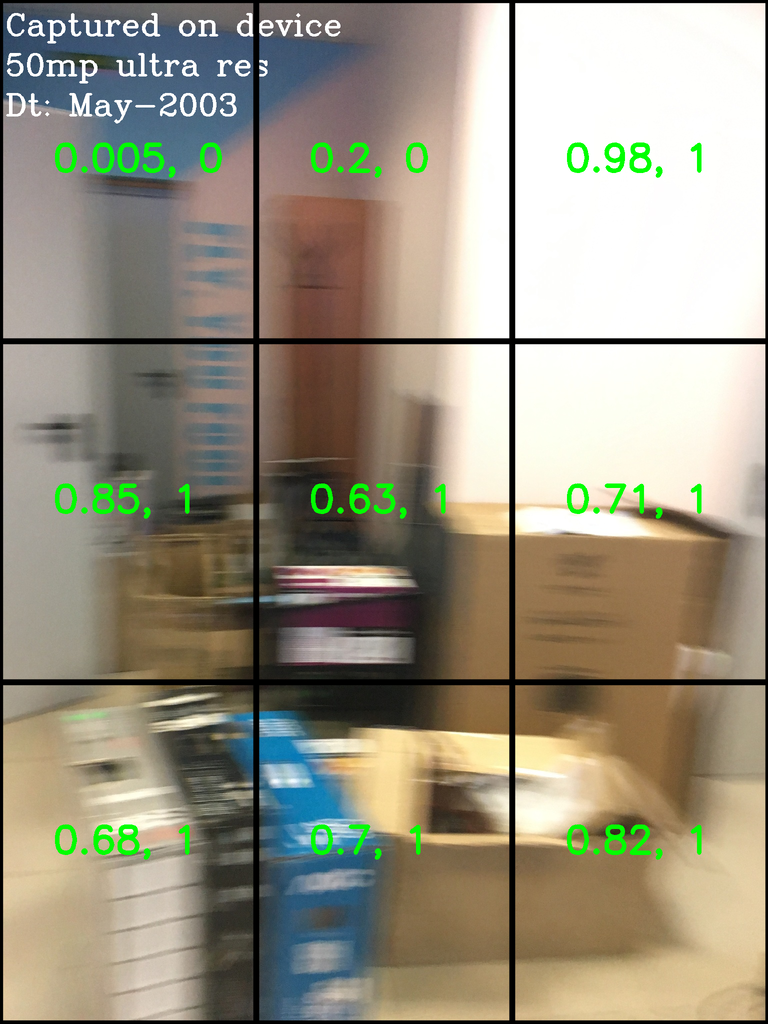}\label{watermark-grid}}

\end{tabular}
}
\caption{
Blurred image from KBD. \textbf{(a)} is the original image and \textbf{(b)} is the same image with a text watermark on the top left. Adding a watermark we observed a 10.7\% increase in the mean of the laplacian map, along with an increase of 39.8\% in the standard-deviation. Similarly we also observed a 9.5\% and 30.2\% increase respectively in the mean and standard deviation of the tenengrad map. In \textbf{(c)} we apply patch-wise classification and plot the predictions on each image grid, where each patch depicts the probability-score of an image being blur and the blur prediction. In \textbf{(d)} we can see how the text watermark affects the probability score and the decision of the model.
}
\label{fig:watermark-comparisons}
\end{figure*}

\section{Proposed Method}

We propose a method that uses conceptually simple spatial image features, along with their statistical measures such as the mean and the variance. These features rely on the fact that the texture of sharp images is different from that of a blurred image. As the image is subjected to blur the edge details within the scene are reduced. More precisely, the strength and the count of strong edges in blurred images are lower compared to that of a sharp image in other words, in blurred images the gradient follows a heavy tailed distribution. Additionally, blur causes a smoothening effect in images which implies that in a neighbourhood the pixel intensities are closer to the central pixel. The features extracted are as follows:

\begin{enumerate}
    \setlength\itemsep{1em}
    \item \textbf{Normalized gray level variance} This operator is an indication of the overall intensity of the image Eqn. (\ref{eq:norm-glv}). The intensity distribution of a blurred image is relatively packed and on the lower spectrum with an overall uniform distribution of intensity when compared to sharp images. 
    
    \begin{equation} \label{eq:norm-glv}
        NGLV = \frac{\sum_{(i, j) \in \Omega(x, y)}(I_{i, j} - \mu) ^ 2}{\mu(x, y)}
    \end{equation}
    \textit{where $\mu(x,y)$ is the mean value for computed over the neighborhood window  $\Omega(x,y)$}
    \item \textbf{Tenengrad} Sobel is a first order derivative operator and indicates the spatial locations of change in intensity of an image. Locations that tend to have a high change in intensities represent stronger edges. We use the horizontal and vertical Sobel maps to obtain the Tenengrad map Eqn. (\ref{eq:tenengrad-eq}) and use the mean of this operator as our feature.
    
    \begin{equation} \label{eq:tenengrad-eq}
        TEN(x, y) = \sum_{(i, j)\in \Omega(x, y)}(S_{x}(i, j)^2 + S_{y}(i, j) ^ 2) 
    \end{equation}
    
    \textit{where $S_x$ and $S_y$ are the Sobel gradients in x and y direction}
    
    \item \textbf{Laplacian} Laplacian operator is the second-order derivative of the input signal. It is highly sensitive to the noise in the input image when compared to Sobel. Laplacian produces high values where there is a rapid change of intensities. We convolve the input image with a Laplacian operator and obtain the mean of the Laplacian map and also extract the variance from the Laplacian map.
    
    \begin{equation}
        Laplacian_{var}(x, y) = \sum_{(i, j) \in \Omega(x, y)} (\Delta I(i, j) - \Bar{\Delta I}) ^ 2
    \end{equation}
    
    \textit{where $\Bar{\Delta I}$ is the mean value of the Laplacian map in a neighbourhood $\Omega(x, y)$}
    
    \item \textbf{LBP Sharpness Map} In \cite{yi2016lbp} the authors propose that the LBP feature descriptor can be leveraged to accurately detect blur within an image patch. For a blurred image, LBP operator relies on the fact that the intensities in a neighbourhood are closer to the central pixel. We use the statistical mean and variance of the LBP descriptor of an image. These features can even help to distinguish between partially blurred images with intended blur from completely unintended blurred images.
    
    \begin{equation}\label{eq:modified-lbp}
        M_{LBP}(x, y) = \frac{\sum_{i=6}^{9}n(LBP_{8, 1}^{riu2}i)}{N}
    \end{equation}
    
    \textit{where $n(LBP_{8, 1}^{riu2}i)$ represents the number of rotation-invariant uniform 8-bit LBP patterns of type i, and N is the total number of pixels in the neighbourhood}
    
\end{enumerate}

\begin{table}[t]
\centering
\caption{Selected set of features for blur classification. These features are calculated at global or in a patch-wise manner at local level.}
\label{tab:blur-features}
\begin{tabular}{p{0.38\linewidth} p{0.38\linewidth} c}
\toprule
\textbf{Feature}               & \textbf{Description}                               & \textbf{Index} \\ \midrule
\textit{Laplacian}    & Mean and variance of the laplacian map & 1-2 \\ 
\textit{Tenengrad}    & Mean of the tenengrad map                  & 3   \\ 
\textit{Normalized gray level variance} & Combination of variance and mean from the grayscale image & 4 \\ 
\textit{LBP Sharpness Map} & Mean and variance of the sharpness map & 5-6 \\ \bottomrule
\end{tabular}
\end{table}

On the whole, we compute the Tenengrad gradient, Laplacian gradient, LBP sharpness map and obtain statistical measures of these maps and the Normalized gray level variance of the input image and stack these together to create a feature vector (Table \ref{tab:blur-features}). This descriptor is used for classifying an input image as blur or sharp. We compute the features at a global level, on the entire image. Our experiments (refer experiments \ref{experiments}) reveal that global level features fail to capture finer details of the image and this leads to some obvious misclassifications. We improve upon this by employing the proposed features in a patch-wise manner to capture finer context. We train XGBoost and CNN models using these features for the task of blur classification and report the metrics on different datasets.

\section{Dataset}

For the purpose of our experiments, we have used a kaggle dataset\footnote[1]{\href{https://www.kaggle.com/datasets/kwentar/blur-dataset}{Kaggle Blur Dataset - https://www.kaggle.com/datasets/kwentar/blur-dataset}} for training and validation, which we refer to as Kaggle Blur Dataset (KBD). The dataset contains a total of 1050 images split equally across the 3 categories namely sharp, motion-blur, and defocus-blur. Since we do not distinguish between the type of blur in images we have combined the two blur categories into a single category which gives us a total of 700 blur images and 350 sharp images.

To demonstrate the generlizability of our approach, we evaluate our model on the BHBID Dataset from \cite{wang2019automatic}. This dataset contains a total of 1188 sample images. The train split contains 418 blurred images and 200 clear images. While the test split contains the remaining 210 blurred images and 80 sharp images. 

We also evaluate the performance of this model on our internal test set. This dataset contains indoor images captured from mobile cameras. This contains a total of 407 images of varying resolution. Among all the samples a total of 202 images are blurred and the rest are sharp.

\section{Experiments} \label{experiments}

In order to compare the different methods we use the KBD to train and validate all of our models. We use BHBID and our internal dataset to test the trained models. We choose XGBoost for this task as it perfectly fits our need for a fast and accurate classifier. Additionally, we train several CNN networks for the blur classification task. In the subsequent section, we discuss the improvements in feature selection and compare the performance of our experiments in Table \ref{table:metrics}.  

We train and evaluate the classifiers in a Stratified KFold Cross-validation manner with K set to 5 and repeat this for 5 random shuffles of the dataset. For each run, we use 25\% of the samples as our validation set and 75\% of the data is used to train our models. We train the XGBoost model with its parameters (max-depth, learning-rate, n-estimators, gamma) set to default values of (6, 0.3, 100, 0) and use the default binary:logistic (binary cross-entropy) as our objective function. 

\subsection{Global Image Features} \label{global-image-features}
In the feature extraction stage, spatial image features - tenengrad mean, laplacian mean and variance, and normalized gray level variance are extracted from grayscale images. With this 4-dimensional feature vector as input, we train a XGBoost model. This model achieves a mean AUC of 0.933 and a mean accuracy of 0.869. On our internal test set we observed missclassification of some blurred images, this was due to a text watermark that is introduced by some smartphones while capturing images. Such artefacts skew the gradients and global features especially mean and variance cannot encapsulate these local features Fig.\ref{fig:watermark-comparisons}. 

\begin{figure*}[t!]
    \centering
    \begin{tabular}{cc}
    \subfloat[Mean accuracy]{\includegraphics[width=.48\linewidth]{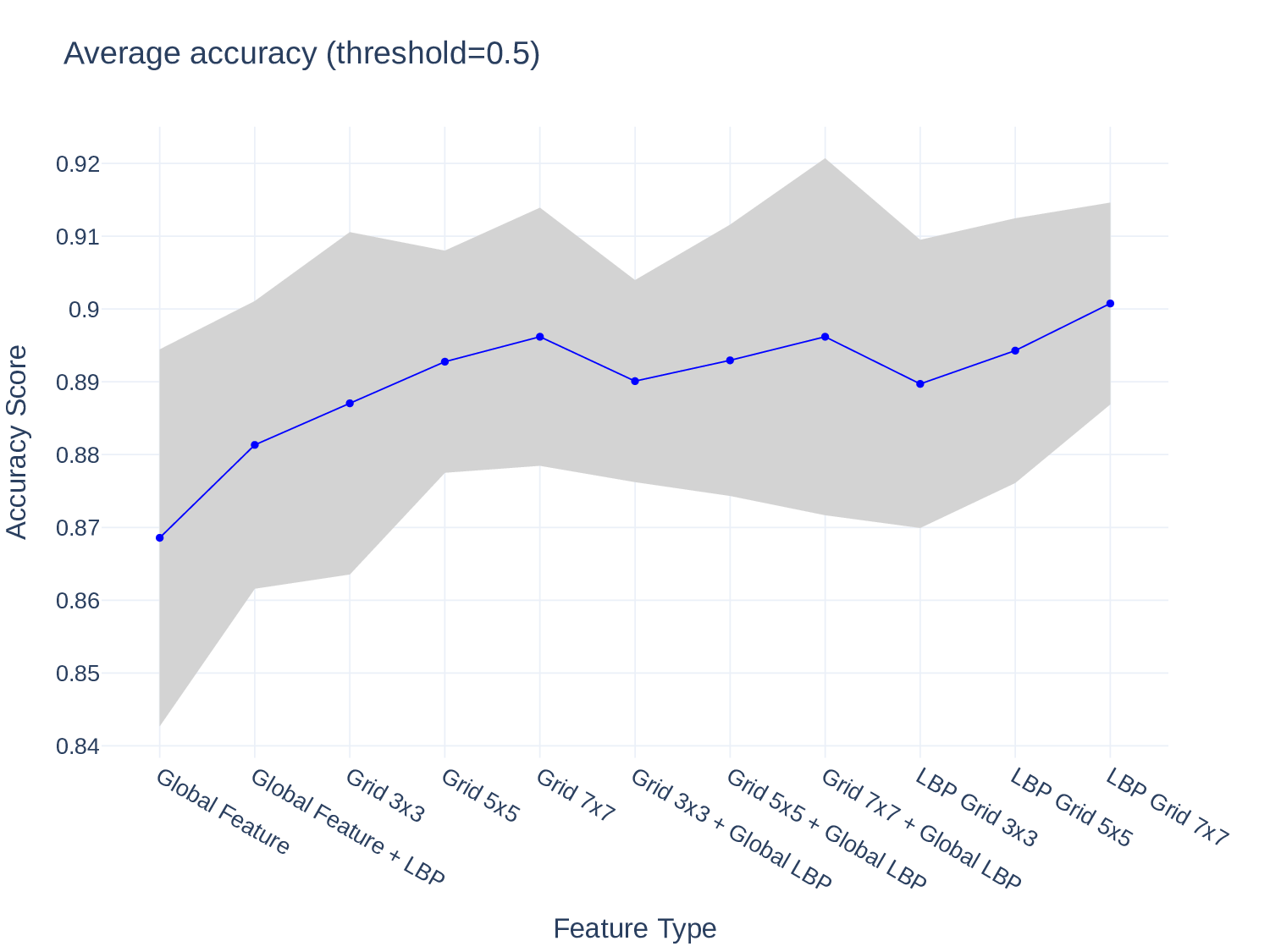}}
    \subfloat[Mean AUC]{\includegraphics[width=.48\linewidth]{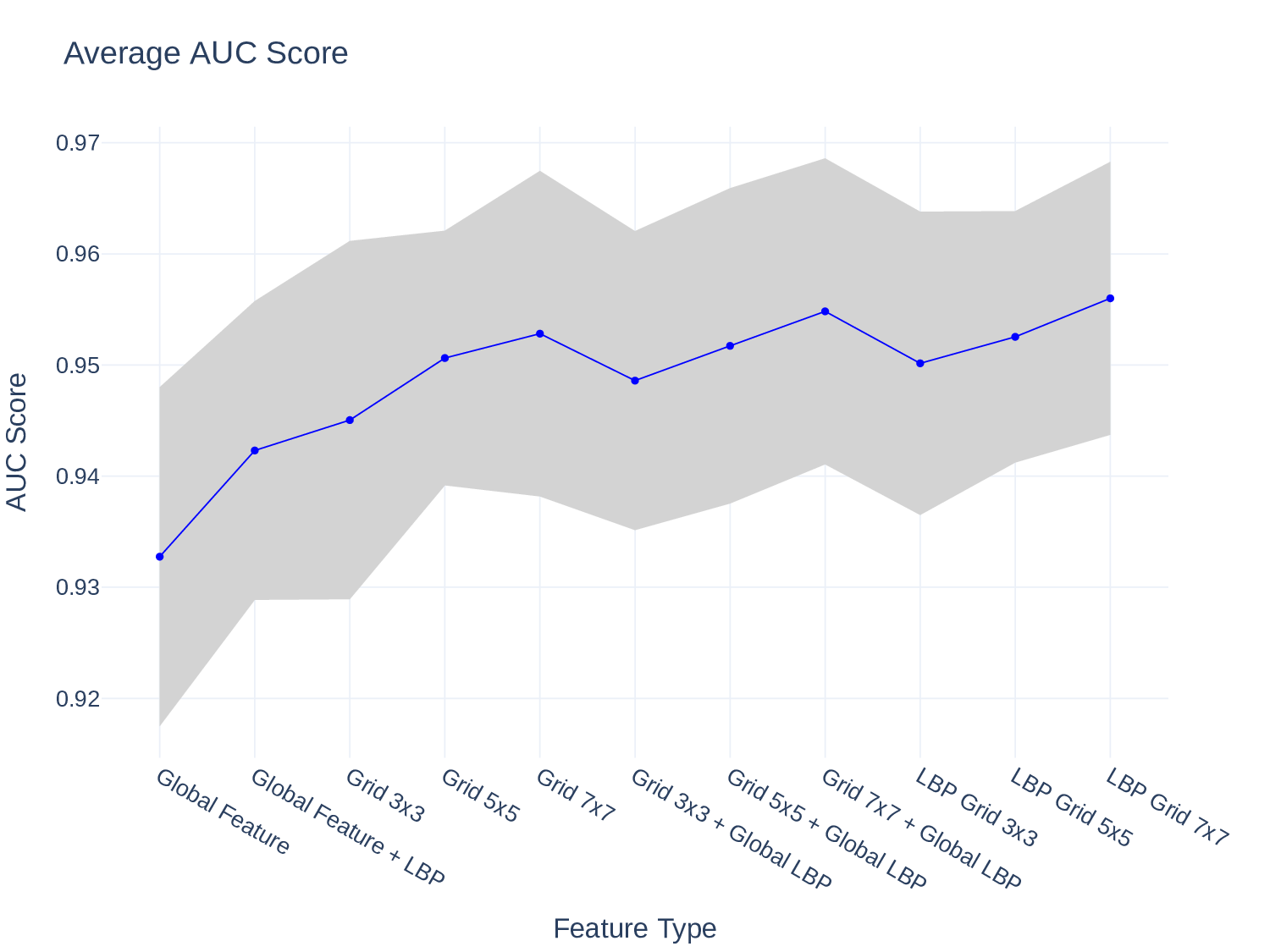}}
    \end{tabular}
    \caption{Figure shows the mean accuracy and mean auc plots for the XGB models trained on different features, it can be observed that the metrics increase as we move to finer and more detailed features}
    \label{fig:xgb-combined-metrics}
\end{figure*}

\subsection{Patch-wise Image Features} \label{patch-image-features}
To investigate the discriminative ability of the features, we isolated the misclassified images, and conduct experiments by cropping the image regions with text. The models classify the watermarked images accurately in this case. Based on this we develop a patch-wise voting mechanism for blur classification. We can observe from Figure \ref{fig:watermark-comparisons} that the watermarks cause a degradation in prediction confidence. 

Rather than employing our classifier in a sliding window fashion on each patch within a single image, we split the image into 3x3 regular size grids and extract features from each of the grid in a single pass. The features are concatenated to produce single feature vector for each image. We refer to these as patch-wise features. With these features, we observe an increase of 1.8\% in accuracy and an increase of 1.2 units in AUC. 

We conduct ablation on the number of patches, by increasing in number of grids within an image from 3x3 to 5x5 and 7x7 grids. For each we train XGBoost classifiers, and observe an average increase of 2.3\% in mean acc and an average increase of 1.67 units in mean auc compared to global features as we move to \textit{smaller grid sizes and a larger number of grids}. This is because high number of patches capture details from the images on a much finer level. We observe that for high resolution images a large number of grids at 7x7 patches are well suited, while for lower resolution images, a smaller number of grids at 3x3 patches perform better. The best results were observed when we use patch-wise features extracted on 7x7 grids, we refer to this as \textit{Grid 7x7} in Table \ref{table:metrics}. 

\subsection{LBP Features} \label{lbp-features}
We investigate the addition of the global mean and variance of the LBP features to our feature descriptor, resulting in a 6 dimensional feature vector. Training the model on these features we observed a mean increase of 1 unit in AUC and a mean increase of 1.2\% in accuracy. 

We further conduct ablation on local patch-wise features by adding global LBP features and patch-wise LBP features. We observe similar trends (Figure \ref{fig:xgb-combined-metrics}) in the increase of AUC and accuracy. We achieve the best performance, an average AUC of 0.956 and average accuracy of 90.1\%, using \textit{LBP Grid 7x7} features. \par
We compare all these methods on the validation sets, plot the metrics in Figure \ref{fig:xgb-combined-metrics} and report them for comparison in Table \ref{table:metrics}

\subsection{Comparison with convolutional neural networks} \label{cnn-intro}

To present an effective benchmark of performance, we compare our proposed features with various CNNs. We train two CNN classifiers which have a vgg16 backbone. These networks were fine-tuned from different tasks, one was used as an encoder in a defocus segmentation task, we refer to this as \textit{vgg-defocus} and the other was trained on imagenet for classification, we refer to this as \textit{vgg-imagenet}. VGG16 is a heavy and memory hungry network that is orders of magnitude complex compared to our handcrafted features, with this complexity in mind, we also train a simple cnn network with 6 layers (4 conv + 2 linear). We also fine-tune a Mobilenet \cite{sandler2018mobilenetv2} network which is preferable for low-latency and low-power systems. We compare these over evaluation criteria such as AUC, accuracy.

\begin{table}[t]
    \centering
    \caption{Improvements observed by introducing more features and local features. We report the average accuracy and auc score of different the methods across all the validation sets of the KFold cross validation. For the CNN networks we report metrics on input resolution and downsampled images to 224x224.}
    \label{table:metrics}
    \centering
    \resizebox{0.475\textwidth}{!}{
      \begin{tabular}{|cc|c|c|}
        \hline
        \multicolumn{2}{|c|}{\textbf{Feature Type}} & Accuracy & AUC \\ \hline
        \multicolumn{1}{|c|}{\multirow{2}{*}{\textbf{Global Features}}} & Global Feature & $86.9 \pm 3$ & $0.933 \pm 0.02$ \\ \cline{2-4}
        \multicolumn{1}{|c|}{} & Global Feature + LBP & $88.1 \pm 2$ & $0.943 \pm 0.01$ \\ \hline
        \multicolumn{1}{|c|}{\multirow{3}{*}{\textbf{Patch-wise Features (No LBP)}}} & Grid 3x3 & $88.7 \pm 2 $ & $0.945 \pm 0.02$ \\ \cline{2-4}
        \multicolumn{1}{|c|}{} & Grid 5x5 & $89.3 \pm 2$ & $0.951 \pm 0.01$ \\ \cline{2-4}
        \multicolumn{1}{|c|}{} & Grid 7x7 & $89.6 \pm 1$ & $0.953 \pm 0.01$ \\ \hline
        \multicolumn{1}{|c|}{\multirow{3}{*}{\textbf{Patch-wise Features + Global LBP}}} & Grid 3x3 + Global LBP & $89 \pm 1$   & $0.949 \pm 0.01$ \\ \cline{2-4}
        \multicolumn{1}{|c|}{} & Grid 5x5 + Global LBP & $89.3 \pm 2$ & $0.952 \pm 0.01$ \\ \cline{2-4}
        \multicolumn{1}{|c|}{} & Grid 7x7 + Global LBP & $89.6 \pm 2$ & $0.955 \pm 0.01$ \\ \hline
        \multicolumn{1}{|c|}{\multirow{3}{*}{\textbf{Patch-wise Features}}} & LBP Grid 3x3 & $89 \pm 2 $ & $0.95 \pm 0.01 $ \\ \cline{2-4}
        \multicolumn{1}{|c|}{} & LBP Grid 5x5 & $89.4 \pm 2$ & $0.953 \pm 0.01$ \\ \cline{2-4}
        \multicolumn{1}{|c|}{} & \textbf{LBP Grid 7x7} & $\boldsymbol{90.1 \pm 1}$ & $\boldsymbol{0.956 \pm 0.01}$ \\ \hline
        \multicolumn{1}{|c|}{\multirow{4}{*}{\textbf{CNN Methods}}} & $\boldsymbol{vgg-defocus^\ast}$ & $\boldsymbol{94.2 \pm 1.6}$ & $\boldsymbol{0.985 \pm 0.01}$ \\ \cline{2-4}
        \multicolumn{1}{|c|}{} & $vgg-imagenet^\ast$ & $93.1 \pm 1.6$ & $\boldsymbol{0.971 \pm 0.02}$ \\ \cline{2-4}
        \multicolumn{1}{|c|}{} & vgg-defocus & $83.2 \pm 2$ & $0.894 \pm 0.04$ \\ \cline{2-4}
        \multicolumn{1}{|c|}{} & vgg-imagenet & $78.3 \pm 2$ & $0.883 \pm 0.04$ \\ \hline
      \end{tabular}}
    {\raggedright $\ast$ The metrics were calculated on downscaled images}
\end{table}

\begin{figure*}[!ht]
    \centering
    \subfloat{\includegraphics[width=.48\linewidth]{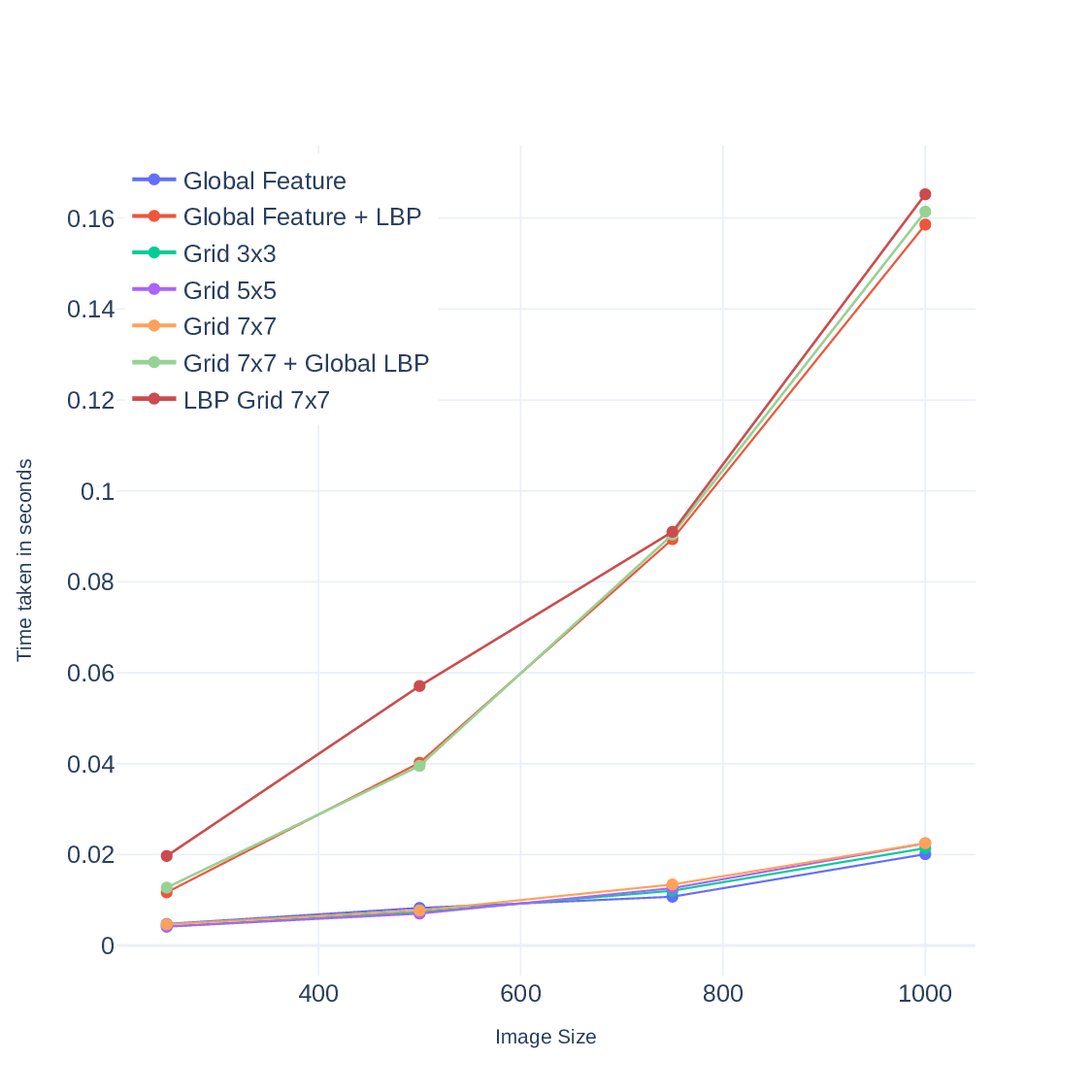}} 
    \subfloat{\includegraphics[width=.48\linewidth]{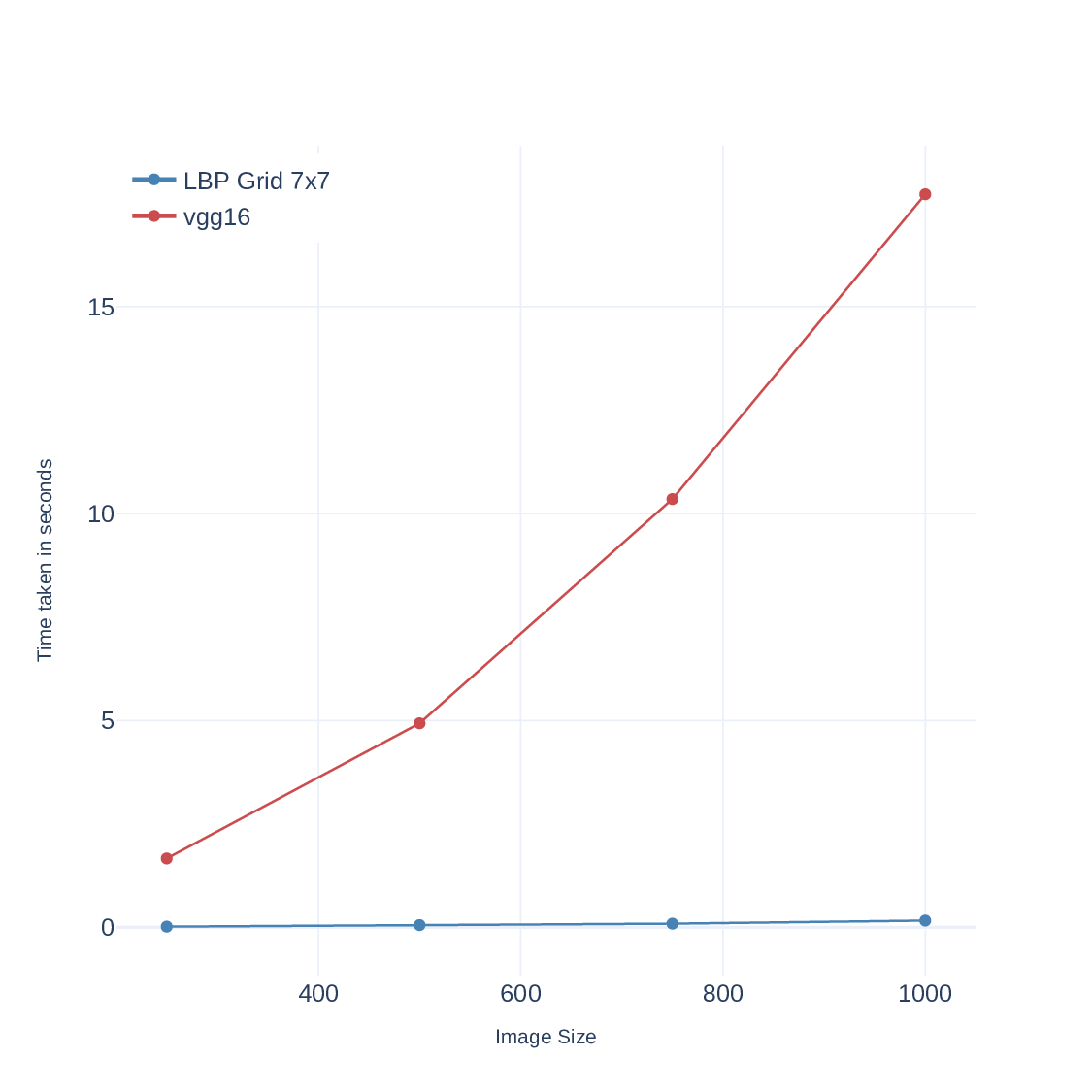}}
    \caption{Comparison of runtimes across different methods. We take 5 different images from KBD, resize it to different sizes and record the time required to run inference on these. It can be seen that the best performing feature \textit{LBP Grid 7x7} is 10 times faster than VGG}
    \label{fig:timing-benchmark}
\end{figure*}

KBD contains images of varying resolution. In order to train the networks, we follow the method laid out in~\cite{krizhevsky2012imagenet} and resize the shorter side of the image to 224 while preserving the aspect ratio. We then perform a center crop of 224x224 from the resized image. For fine-tuning, we freeze the convolutional layers in the networks and modify the structure of the final layers for binary classification. We use data augmentation, which includes a random horizontal and random vertical flip with a probability of 0.6. We fine tune the network using an Adam optimizer with a learning rate set to 1e-3, a batch size of 256, and binary cross entropy loss as our objective function. Similar to the XGBoost setup, we train these networks in a Stratified KFold Cross Validation manner with 25\% of the data being used as validation set in each run. To prevent overfitting, we train the vgg networks for 5 epochs each and the other networks for 10 epochs each.

We report metrics on the original image resolution and also on images downscaled to 224x224. The adaptive average pooling layer handles the variable input size in the case of original image resolution. We report the mean AUC and mean accuracy metrics on the validation splits in Table \ref{table:metrics}. 

\begin{table}[ht]
\centering
\caption{Comparison of CNN methods - we measure the inference time on images of size 1000x1000}
\label{tab:cnn-metrics}
\begin{tabular}{p{0.35\linewidth} p{0.15\linewidth} p{0.16\linewidth} r}
\toprule
\textbf{Network} & \textbf{Param Count}  & \textbf{Inference Time} & \textbf{Accuracy} \\ \midrule
\textit{VGG16 $\ast$}    & 138M & 16.4s & 93.1 \\ 
\textit{MobileNetv2 $\ast$}    & 2.3M & 4.3s & 90.8 \\ 
\textit{Simple CNN Classifier $\ast$} & 0.96M & 0.94s & 80.5 \\ 
\bottomrule
\end{tabular}
{\raggedright $\ast$ Accuracy calculated on downscaled images}
\end{table}

\subsection{Results}
We benchmark the inference time required to classify an image across all the methods. All the models are run on a computer with an Intel i5-8250U CPU running at 1800 MHz using 8GB RAM, running Ubuntu 20.04.3 LTS and Python 3.8.10, no other compiler optimizations are used. For the XGBoost model, we measure the time required to extract the features and perform the classification. For vgg16 we set the batch size to 1 and record the time elapsed for the forward pass. For all the algorithms, we measure the time required to process 5 different images, we repeat this experiment for 10 runs and report the mean of the 10 runs in Fig. \ref{fig:timing-benchmark}.

We also test our extracted features using support vector machines (SVM) classifiers and compare the performance to XGBoost classifier. They perform 3, 3.4, 3.4 percentage points (accuracy) lower than XGBoost in case of Grid 7x7, Grid 7x7 + Global LBP, LBP Grid 7x7. This shows that the extracted features are discriminative across various classifiers. 

\begin{table}[t]
\centering
\caption{Performance of our best feature set (LBP Grid 7x7) on two different test sets.}
\begin{tabular}{l|c|c|c|c}
\hline
\multirow{2}{*}{Dataset} &
\multirow{2}{*}{Accuracy} &
\multirow{2}{*}{AUC} &
\multicolumn{2}{c}{F1 Score} \\
\cline{4-5}
 & & & Blur & Sharp \\
\hline
\hline
BHBID $(Test Set)$ & 98  & 0.98  & 0.98  & 0.96 \\ 
Internal Dataset & 91  & 0.96 & 0.89  & 0.91 \\ 
\hline
\end{tabular}
\label{tab:test-metrics}
\end{table}

When discussing CNN classifiers, it is worth noting that the VGG network provides the best performance but is very large and requires a lot of memory. In contrast, the lightweight Mobilenet is reasonably fast and accurate but does not compare well in speed with our selected features. Simple CNN is very fast but it performs worse than Mobilenet Table \ref{tab:cnn-metrics} 

To demonstrate the generalization of our method we test our methods on two additional datasets. We train our model on the training split of BHBID Dataset and report the results achieved by our best model \textit{LBP Grid 7x7}. We also report the metrics on our internal test set, we use the XGBoost model that was trained on KBD out-of-the-box and perform classification using our best model \textit{LBP Grid 7x7}. The results are summarized in Table
\ref{tab:test-metrics}.

A drawback of this approach is that this fails when none of the patches in the image contain features in other words a plain image with no noticeable difference in intensity, in such cases the images are classified as blur.

\section{Conclusion}
In this work, we address the task of blur classification for image quality assessment. We extract various spatial and statistical image features to classify an input image as blurred or sharp. And empirically demonstrate how extracting features at a global level fails to capture the intricate details in an image. We propose a patch-wise method for feature extraction and show its effectiveness for blur classification on multiple datasets. We also apply this method to our internal dataset without any further tweaks after training on KBD. We train XGBoost models and show the superiority of our features both in terms of inference time and classification metrics. Our best performing feature is the LBP Grid 7x7 that has an AUC of \textbf{0.95} on KBD and an AUC of \textbf{0.98} on the BHBID Dataset. We compare our features against pretrained as well as lightweight CNN models, and find that they are 6.9 percentage points better than the VGG model (at original image resolution) and are twice as fast when compared to the low-latency Mobilenet. 

\bibliographystyle{ACM-Reference-Format}
\bibliography{pbic-source}


\begin{thebibliography}{17}


\ifx \showCODEN    \undefined \def \showCODEN     #1{\unskip}     \fi
\ifx \showDOI      \undefined \def \showDOI       #1{#1}\fi
\ifx \showISBNx    \undefined \def \showISBNx     #1{\unskip}     \fi
\ifx \showISBNxiii \undefined \def \showISBNxiii  #1{\unskip}     \fi
\ifx \showISSN     \undefined \def \showISSN      #1{\unskip}     \fi
\ifx \showLCCN     \undefined \def \showLCCN      #1{\unskip}     \fi
\ifx \shownote     \undefined \def \shownote      #1{#1}          \fi
\ifx \showarticletitle \undefined \def \showarticletitle #1{#1}   \fi
\ifx \showURL      \undefined \def \showURL       {\relax}        \fi
\providecommand\bibfield[2]{#2}
\providecommand\bibinfo[2]{#2}
\providecommand\natexlab[1]{#1}
\providecommand\showeprint[2][]{arXiv:#2}

\bibitem[Ali and Mahmood(2018)]%
        {ali2018analysis}
\bibfield{author}{\bibinfo{person}{Usman Ali} {and}
  \bibinfo{person}{Muhammad~Tariq Mahmood}.} \bibinfo{year}{2018}\natexlab{}.
\newblock \showarticletitle{Analysis of blur measure operators for single image
  blur segmentation}.
\newblock \bibinfo{journal}{\emph{Applied Sciences}} \bibinfo{volume}{8},
  \bibinfo{number}{5} (\bibinfo{year}{2018}), \bibinfo{pages}{807}.
\newblock


\bibitem[Basha et~al\mbox{.}(2018)]%
        {basha2018rccnet}
\bibfield{author}{\bibinfo{person}{SH~Shabbeer Basha}, \bibinfo{person}{Soumen
  Ghosh}, \bibinfo{person}{Kancharagunta~Kishan Babu},
  \bibinfo{person}{Shiv~Ram Dubey}, \bibinfo{person}{Viswanath Pulabaigari},
  {and} \bibinfo{person}{Snehasis Mukherjee}.} \bibinfo{year}{2018}\natexlab{}.
\newblock \showarticletitle{Rccnet: An efficient convolutional neural network
  for histological routine colon cancer nuclei classification}. In
  \bibinfo{booktitle}{\emph{2018 15th International Conference on Control,
  Automation, Robotics and Vision (ICARCV)}}. IEEE,
  \bibinfo{pages}{1222--1227}.
\newblock


\bibitem[Bojarski et~al\mbox{.}(2016)]%
        {bojarski2016end}
\bibfield{author}{\bibinfo{person}{Mariusz Bojarski}, \bibinfo{person}{Davide
  Del~Testa}, \bibinfo{person}{Daniel Dworakowski}, \bibinfo{person}{Bernhard
  Firner}, \bibinfo{person}{Beat Flepp}, \bibinfo{person}{Prasoon Goyal},
  \bibinfo{person}{Lawrence~D Jackel}, \bibinfo{person}{Mathew Monfort},
  \bibinfo{person}{Urs Muller}, \bibinfo{person}{Jiakai Zhang},
  {et~al\mbox{.}}} \bibinfo{year}{2016}\natexlab{}.
\newblock \showarticletitle{End to end learning for self-driving cars}.
\newblock \bibinfo{journal}{\emph{arXiv preprint arXiv:1604.07316}}
  (\bibinfo{year}{2016}).
\newblock


\bibitem[Cun and Pun(2020)]%
        {cun2020defocus}
\bibfield{author}{\bibinfo{person}{Xiaodong Cun} {and} \bibinfo{person}{Chi-Man
  Pun}.} \bibinfo{year}{2020}\natexlab{}.
\newblock \showarticletitle{Defocus blur detection via depth distillation}. In
  \bibinfo{booktitle}{\emph{European Conference on Computer Vision}}. Springer,
  \bibinfo{pages}{747--763}.
\newblock


\bibitem[He et~al\mbox{.}(2016)]%
        {he2016deep}
\bibfield{author}{\bibinfo{person}{Kaiming He}, \bibinfo{person}{Xiangyu
  Zhang}, \bibinfo{person}{Shaoqing Ren}, {and} \bibinfo{person}{Jian Sun}.}
  \bibinfo{year}{2016}\natexlab{}.
\newblock \showarticletitle{Deep residual learning for image recognition}. In
  \bibinfo{booktitle}{\emph{Proceedings of the IEEE conference on computer
  vision and pattern recognition}}. \bibinfo{pages}{770--778}.
\newblock


\bibitem[Krizhevsky et~al\mbox{.}(2012)]%
        {krizhevsky2012imagenet}
\bibfield{author}{\bibinfo{person}{Alex Krizhevsky}, \bibinfo{person}{Ilya
  Sutskever}, {and} \bibinfo{person}{Geoffrey~E Hinton}.}
  \bibinfo{year}{2012}\natexlab{}.
\newblock \showarticletitle{Imagenet classification with deep convolutional
  neural networks}.
\newblock \bibinfo{journal}{\emph{Advances in neural information processing
  systems}}  \bibinfo{volume}{25} (\bibinfo{year}{2012}).
\newblock


\bibitem[Kupyn et~al\mbox{.}(2019)]%
        {kupyn2019deblurgan}
\bibfield{author}{\bibinfo{person}{Orest Kupyn}, \bibinfo{person}{Tetiana
  Martyniuk}, \bibinfo{person}{Junru Wu}, {and} \bibinfo{person}{Zhangyang
  Wang}.} \bibinfo{year}{2019}\natexlab{}.
\newblock \showarticletitle{Deblurgan-v2: Deblurring (orders-of-magnitude)
  faster and better}. In \bibinfo{booktitle}{\emph{Proceedings of the IEEE/CVF
  International Conference on Computer Vision}}. \bibinfo{pages}{8878--8887}.
\newblock


\bibitem[Liang et~al\mbox{.}(2021)]%
        {liang2021pruning}
\bibfield{author}{\bibinfo{person}{Tailin Liang}, \bibinfo{person}{John
  Glossner}, \bibinfo{person}{Lei Wang}, \bibinfo{person}{Shaobo Shi}, {and}
  \bibinfo{person}{Xiaotong Zhang}.} \bibinfo{year}{2021}\natexlab{}.
\newblock \showarticletitle{Pruning and quantization for deep neural network
  acceleration: A survey}.
\newblock \bibinfo{journal}{\emph{Neurocomputing}}  \bibinfo{volume}{461}
  (\bibinfo{year}{2021}), \bibinfo{pages}{370--403}.
\newblock


\bibitem[Masi et~al\mbox{.}(2018)]%
        {masi2018deep}
\bibfield{author}{\bibinfo{person}{Iacopo Masi}, \bibinfo{person}{Yue Wu},
  \bibinfo{person}{Tal Hassner}, {and} \bibinfo{person}{Prem Natarajan}.}
  \bibinfo{year}{2018}\natexlab{}.
\newblock \showarticletitle{Deep face recognition: A survey}. In
  \bibinfo{booktitle}{\emph{2018 31st SIBGRAPI conference on graphics, patterns
  and images (SIBGRAPI)}}. IEEE, \bibinfo{pages}{471--478}.
\newblock


\bibitem[Ojala et~al\mbox{.}(2002)]%
        {ojala2002multiresolution}
\bibfield{author}{\bibinfo{person}{Timo Ojala}, \bibinfo{person}{Matti
  Pietikainen}, {and} \bibinfo{person}{Topi Maenpaa}.}
  \bibinfo{year}{2002}\natexlab{}.
\newblock \showarticletitle{Multiresolution gray-scale and rotation invariant
  texture classification with local binary patterns}.
\newblock \bibinfo{journal}{\emph{IEEE Transactions on pattern analysis and
  machine intelligence}} \bibinfo{volume}{24}, \bibinfo{number}{7}
  (\bibinfo{year}{2002}), \bibinfo{pages}{971--987}.
\newblock


\bibitem[Sandler et~al\mbox{.}(2018)]%
        {sandler2018mobilenetv2}
\bibfield{author}{\bibinfo{person}{Mark Sandler}, \bibinfo{person}{Andrew
  Howard}, \bibinfo{person}{Menglong Zhu}, \bibinfo{person}{Andrey Zhmoginov},
  {and} \bibinfo{person}{Liang-Chieh Chen}.} \bibinfo{year}{2018}\natexlab{}.
\newblock \showarticletitle{Mobilenetv2: Inverted residuals and linear
  bottlenecks}. In \bibinfo{booktitle}{\emph{Proceedings of the IEEE conference
  on computer vision and pattern recognition}}. \bibinfo{pages}{4510--4520}.
\newblock


\bibitem[Sankhe et~al\mbox{.}(2011)]%
        {sankhe2011deblurring}
\bibfield{author}{\bibinfo{person}{PD Sankhe}, \bibinfo{person}{M Patil}, {and}
  \bibinfo{person}{M Margaret}.} \bibinfo{year}{2011}\natexlab{}.
\newblock \showarticletitle{Deblurring of grayscale images using inverse and
  Wiener filter}. In \bibinfo{booktitle}{\emph{Proceedings of the International
  Conference \& Workshop on Emerging Trends in Technology}}.
  \bibinfo{pages}{145--148}.
\newblock


\bibitem[Shi et~al\mbox{.}(2014)]%
        {shi2014discriminative}
\bibfield{author}{\bibinfo{person}{Jianping Shi}, \bibinfo{person}{Li Xu},
  {and} \bibinfo{person}{Jiaya Jia}.} \bibinfo{year}{2014}\natexlab{}.
\newblock \showarticletitle{Discriminative blur detection features}. In
  \bibinfo{booktitle}{\emph{Proceedings of the IEEE Conference on Computer
  Vision and Pattern Recognition}}. \bibinfo{pages}{2965--2972}.
\newblock


\bibitem[Szanda{\l}a(2020)]%
        {szandala2020convolutional}
\bibfield{author}{\bibinfo{person}{Tomasz Szanda{\l}a}.}
  \bibinfo{year}{2020}\natexlab{}.
\newblock \showarticletitle{Convolutional neural network for blur images
  detection as an alternative for Laplacian method}. In
  \bibinfo{booktitle}{\emph{2020 IEEE Symposium Series on Computational
  Intelligence (SSCI)}}. IEEE, \bibinfo{pages}{2901--2904}.
\newblock


\bibitem[Vasiljevic et~al\mbox{.}(2016)]%
        {vasiljevic2016examining}
\bibfield{author}{\bibinfo{person}{Igor Vasiljevic}, \bibinfo{person}{Ayan
  Chakrabarti}, {and} \bibinfo{person}{Gregory Shakhnarovich}.}
  \bibinfo{year}{2016}\natexlab{}.
\newblock \showarticletitle{Examining the impact of blur on recognition by
  convolutional networks}.
\newblock \bibinfo{journal}{\emph{arXiv preprint arXiv:1611.05760}}
  (\bibinfo{year}{2016}).
\newblock


\bibitem[Wang et~al\mbox{.}(2019)]%
        {wang2019automatic}
\bibfield{author}{\bibinfo{person}{Rui Wang}, \bibinfo{person}{Wei Li},
  \bibinfo{person}{Rui Li}, {and} \bibinfo{person}{Liang Zhang}.}
  \bibinfo{year}{2019}\natexlab{}.
\newblock \showarticletitle{Automatic blur type classification via ensemble
  SVM}.
\newblock \bibinfo{journal}{\emph{Signal processing: image communication}}
  \bibinfo{volume}{71} (\bibinfo{year}{2019}), \bibinfo{pages}{24--35}.
\newblock


\bibitem[Yi and Eramian(2016)]%
        {yi2016lbp}
\bibfield{author}{\bibinfo{person}{Xin Yi} {and} \bibinfo{person}{Mark
  Eramian}.} \bibinfo{year}{2016}\natexlab{}.
\newblock \showarticletitle{LBP-based segmentation of defocus blur}.
\newblock \bibinfo{journal}{\emph{IEEE transactions on image processing}}
  \bibinfo{volume}{25}, \bibinfo{number}{4} (\bibinfo{year}{2016}),
  \bibinfo{pages}{1626--1638}.
\newblock


\end{thebibliography}

\end{document}